\documentclass{article}

\usepackage{arxiv}

\usepackage[utf8]{inputenc} 
\usepackage[T1]{fontenc}    
\usepackage{hyperref}       
\usepackage{url}            
\usepackage{booktabs}       
\usepackage{amsfonts}       
\usepackage{nicefrac}       
\usepackage{microtype}      
\usepackage{lipsum}
\usepackage{graphicx}
\graphicspath{ {./images/} }
\usepackage{multirow}%
\usepackage{amsmath,amssymb,amsfonts}%
\usepackage{amsthm}%
\usepackage{mathrsfs}%
\usepackage[title]{appendix}%
\usepackage{xcolor}%
\usepackage{textcomp}%
\usepackage{manyfoot}%
\usepackage{booktabs}%
\usepackage{algorithm}%
\usepackage{algorithmicx}%
\usepackage{algpseudocode}%
\usepackage{listings}%

\colorlet{mypurple}{blue!40!pink}
\colorlet{mygreen}{blue!15!green}
\colorlet{myorange}{magenta!45!orange}

\title{Transformers in Time-series Analysis: A Tutorial}

\author{
 Sabeen Ahmed \\
  Department of Machine Learning\\
  Moffitt Cancer Center\\
  12902 USF Magnolia Drive, Tampa, FL, 33612 \\
  \texttt{sabeen.ahmed@moffitt.org} \\
   \And
 Ian E. Nielsen \\
  Department of Electrical and Computer Engineering\\
  Rowan University\\
  201 Mullica Hill Rd, Glassboro, NJ, 08028 \\
  \texttt{nielseni6@rowan.edu} \\
   \And
 Aakash Tripathi \\
  Department of Machine Learning\\
  Moffitt Cancer Center\\
  12902 USF Magnolia Drive, Tampa, FL, 33612 \\
  \texttt{aakash.tripathi@moffitt.org} \\
  \And
 Shamoon Siddiqui \\
  Department of Electrical and Computer Engineering\\
  Rowan University\\
  201 Mullica Hill Rd, Glassboro, NJ, 08028 \\
  \texttt{siddiq76@rowan.edu} \\
   \And
 Ravi P. Ramachandran \\
  Department of Electrical and Computer Engineering\\
  Rowan University\\
  201 Mullica Hill Rd, Glassboro, NJ, 08028 \\
  \texttt{ravi@rowan.edu} \\
   \And
 Ghulam Rasool \\
  Department of Machine Learning\\
  Moffitt Cancer Center\\
  12902 USF Magnolia Drive, Tampa, FL, 33612 \\
  \texttt{ghulam.rasool@moffitt.org} \\
}

\begin{document}
\maketitle
\begin{abstract}
Transformer architectures have widespread applications, particularly in Natural Language Processing and Computer Vision. Recently, Transformers have been employed in various aspects of time-series analysis. This tutorial provides an overview of the Transformer architecture, its applications, and a collection of examples from recent research in time-series analysis. We delve into an explanation of the core components of the Transformer, including the self-attention mechanism, positional encoding, multi-head, and encoder/decoder. Several enhancements to the initial Transformer architecture are highlighted to tackle time-series tasks. The tutorial also provides best practices and techniques to overcome the challenge of effectively training Transformers for time-series analysis.
\end{abstract}


\section{Introduction} \label{sec1}
Transformers belong to a class of machine learning models that use self-attention or the scaled dot-product operation as their primary learning mechanism. Transformers were initially proposed for neural machine translation - one of the most challenging natural language processing (NLP) tasks \cite{Vaswani2017}. Recently, Transformers have been successfully employed to tackle various problems in machine learning and achieve state-of-the-art performance \cite{LIN2022111}. Apart from classical NLP tasks, examples from other areas include image classification \cite{dosovitskiy2020image}, object detection and segmentation \cite{kirillov2023segment}, image and language generation \cite{lu2022unified}, sequential decision-making in reinforcement learning \cite{chen2021decision}, multi-modal (text, speech, and image) data processing \cite{waqas2023multimodal}, and analysis of tabular and time-series data \cite{9414142}. This tutorial paper focuses on time-series analysis using Transformers.

Time-series data consist of ordered samples, observations, or features recorded sequentially over time. Time-series datasets often arise naturally in many real-world applications where data is recorded over a fixed sampling interval. Examples include stock prices, digitized speech signals, traffic measurements, sensor data for weather patterns, biomedical measurements, and various kinds of population data recorded over time. Time-series analysis may include processing the numeric data for multiple tasks, including forecasting, prediction, and classification. Statistical approaches involve using various types of models, such as autoregressive (AR), moving average (MA), auto-regressive moving average (ARMA), AR Integrated MA (ARIMA), and spectral analysis techniques. 

Machine learning models with specialized components and architectures for handling the sequential nature of data have been extensively proposed in the literature and used by the community. The most notable of these machine learning models are Recurrent Neural Networks (RNNs) and their popular variants, including Long Short-Term Memory (LSTM) and Gated Recurrent Units (GRU) \cite{lipton2015critical}, \cite{hochreiter1997long}, \cite{chung2014empirical}, \cite{Dera-TRUST-2023}. These models process batches of data sequentially, one sample at a time, and optimize unknown model parameters using the well-known gradient descent algorithm. The gradient information for updating model parameters is calculated using back-propagation through time (BPTT) \cite{chen2016gentle}. LSTMs and GRUs have been successfully used in many applications \cite{arunkumar2021forecasting}, \cite{che2018recurrent}, \cite{di2017recurrent}, \cite{shewalkar2019performance}, \cite{dixon2021financial}. However, they suffer from several limitations due to the sequential processing of input data and the challenges associated with BPTT, especially while processing datasets with long dependencies. The training process of LSTM and GRU models also suffers from vanishing and exploding gradient problems \cite{fei2018bidirectional}, \cite{ribeiro2020beyond}. While processing long sequences, the gradient descent algorithm (using BPTT) may not update the model parameters as the gradient information is lost (either approaches zero or infinity). Additionally, these models generally do not benefit from the parallel computations offered by graphical processing units (GPUs), tensor processing units (TPUs), and other hardware accelerators \cite{el2021optimized}. Certain architectural modifications and training tricks may help LSTMs and GRUs mitigate gradient-related problems to a certain extent. However, challenges in learning from long data sequences with the limited use of parallelization offered by modern hardware impact the effectiveness and efficiency of RNN-based models \cite{SHERSTINSKY2020132306}.

The Transformer architecture allows for parallel computation of sequential data \cite{tay2022efficient} without a substantial increase in the complexity of the network \cite{Kitaev2020Reformer:}. Transformers, owing to their architecture, can use parallel processing capabilities of Graphics Processing Units (GPUs) and Tensor Processing Units (TPUs) \cite{jouppi2018motivation}. Given the attention-based operations, Transformers can correlate information from all elements in the sequence to each other in parallel without suffering from vanishing gradients as is the case with RNNs and their variants \cite{pascanu2013difficulty},\cite{wang2019learning}, \cite{bapna-etal-2018-training}.

Transformers have substantially improved long-term and multi-variate time-series forecasting \cite{zhang2023crossformer}, \cite{nie2023a}. However, the self-attention mechanism has high computational complexity and memory requirements hampering long sequence modeling. Various modifications have been proposed in the literature to optimize Transformer performance for time-series tasks \cite{liu2021pyraformer}, \cite{zhou2022fedformer}, \cite{chen2022learning}. Training large Transformer models is challenging, especially for massive datasets. Many techniques have been proposed in the literature to efficiently train large Transformer models. These techniques include layer-wise adaptive large batch optimization \cite{You2020Large}, distributed training \cite{huang2019gpipe}, knowledge inheritance \cite{qin2022knowledge}, progressive training \cite{li2022automated}, and mapping parameters of smaller models to initialize larger models \cite{wanglearning}.

The contribution of this tutorial includes,

\begin{itemize}
    \item [1.] Illustrative explanation of the intuition behind the Transformer operations. These intuitions help us understand how Transformers revolutionized NLP tasks.
    \item [2.] Discuss various aspects and techniques introduced in the Transformer architecture and internal operations for efficient time-series analyses.
    \item [3.] Compilation of some use cases of Transformers for time-series analysis, including comparative performance.
    \item [4.] Guidelines on techniques and tricks for efficiently training Transformers.
\end{itemize}


Transformers-based models have revolutionized applications in NLP, computer vision, time-series analysis, and many other areas. However, we have limited our study to time-series applications. Another limitation of the study is related to the variations in the architectures of Transformer models. In recent years, hundreds of variations have been proposed for performance improvement of Transformers. However, we focused on the basic architecture proposed by Vaswani \emph{et al.} \cite{Vaswani2017}.

We start by providing an overview of the Transformer architecture based upon self-attention, scaled dot-product, multi-head, and positional encoding in Section \ref{section:architecture}. Section \ref{section:categories} describes the advancements of Transformers for time-series applications. We then discuss some of the most popular recent time-series Transformer architectures in Section \ref{section:time-series}. Finally, we provide ``best practices'' for training Transformers in Section \ref{section:best-practices} before concluding the paper. 

\section{Transformers: Nuts and Bolts}\label{section:architecture}
We begin by explaining the inner workings of the Transformer as proposed by Vaswani \emph{et al.} in 2017 to solve the challenging problem of neural machine translation \cite{Vaswani2017}. We then proceed with a deep dive into the operations performed inside each component of Transformers and the intuition behind those operations. Several variations of the Transformer architecture have been developed. However, the intuition behind the basic operations used remains the same and is covered in this Section \cite{ho2019axial,roy2021efficient,choromanski2020rethinking}. 
\subsection{The Transformer Architecture}
\label{section:architecturedetails}
The original Transformer is a sequence-to-sequence model designed in an encoder-decoder type of configuration that takes as input a sequence of words from the source language and then generates the translation in the target language \cite{Vaswani2017}. Given that the length of the two sequences and the vocabulary size are not necessarily the same, the model has to learn to encode the source sequence into a fixed-length representation that it can then decode to generate the target sequence in an auto-regressive fashion \cite{bergmeir2018note}. This auto-regressive property comes with a constraint of requiring information to propagate back to the beginning of the sequence during the generation of the translated sequences. The same constraint holds for time-series analysis. 

The machine learning models have been limited by how far back the impact of a specific data sample can be considered during learning. In some cases, the auto-regressive nature of the training of machine learning models leads to memorization of past observations rather than the generalization of the training examples to new data \cite{livska2018memorize,katharopoulos2020transformers}. Transformers address these challenges by using \emph{self-attention} and \emph{positional encoding} techniques to jointly attend to and encode the ordered information as current data samples in the sequence are analyzed. These techniques keep the sequential information intact for learning while eliminating the classical notion of \emph{recurrence} \cite{hao-etal-2019-modeling}. These techniques further allow Transformers to exploit parallelism offered by GPUs and TPUs. Recently, there have been some research attempts to incorporate recurrent components in Transformers \cite{wang2019rtransf}.

\begin{figure}[ht]
\centering
\includegraphics[width=\textwidth]{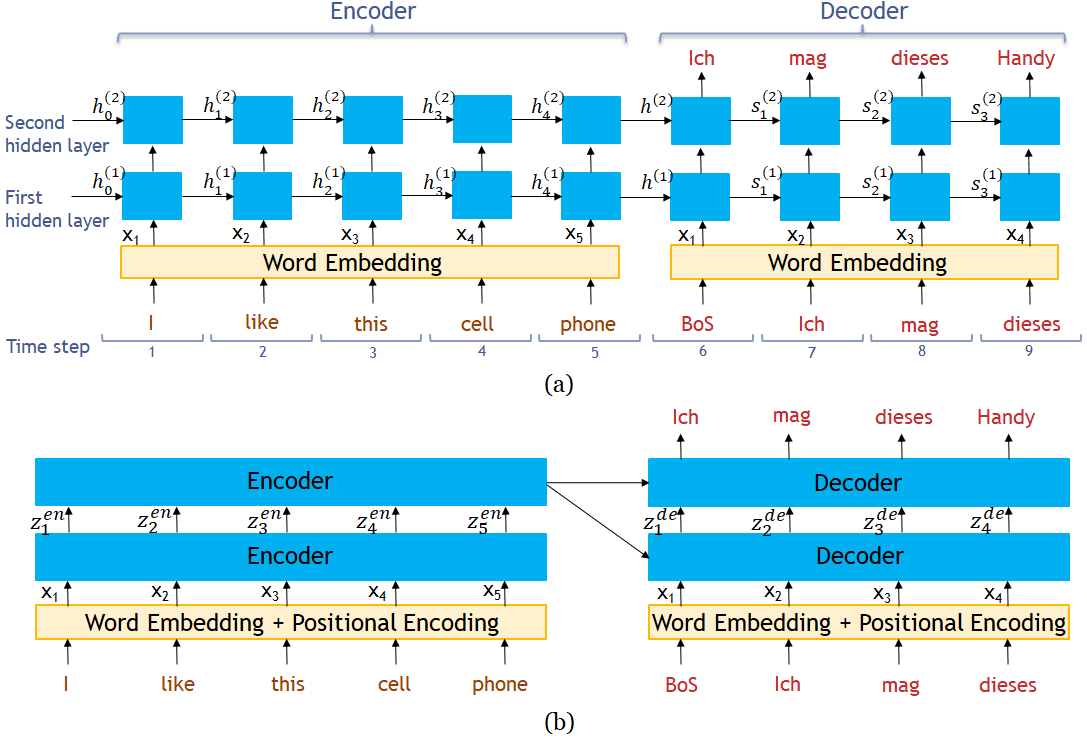}
\caption{An example of a simple language translation task performed using (a) a classical model, such as an RNN, LSTM, or GRU, and (b) a Transformer.}
\label{fig:language-Task-Comparison}
\end{figure}

\textbf{A Simple Translation Example.} Consider an example of translating ``I like this cell phone'' to German ``Ich mag dieses Handy'' using a classical machine translation model (an LSTM or GRU) and a Transformer as illustrated in Figure \ref{fig:language-Task-Comparison}. The input words must be first processed using an embedding layer to convert raw words into vectors of size $d$. The concept of embedding a discrete word into a continuous space of real numbers is a common practice in NLP \cite{mikolov2013efficient,10.1145/1461518.1461544,pennington2014glove,Peters2018DeepRepresentations}. In classical language translation models, each embedded word in a sentence corresponds to a specific RNN/LSTM/GRU cell. The operations in the subsequent cells depend on the output from the previous cell. Therefore, each embedded word in the input is processed sequentially (after processing the preceding word). In a model based on the Transformer architecture, the entire input sequence ``I like this cell phone'' is fed into the model simultaneously, eliminating the need for sequential data processing. The sequence order is tracked using \emph{positional encoding}.

\subsection{Self-Attention Operation}
The Transformer architecture is based on finding associations or relationships between various input segments (after adding the position information in these segments) using the dot product \cite{Gehring2017ConvolutionalLearning}. Let $\{\mathbf{x}_i\}_{i=1}^{n}, \mathbf{x} \in \mathcal{R}^d$ be a set of $n$ words (or data points) in a single sequence. The subscript $i$ represents the position of the vector $\mathbf{x}_i$, equivalently the position of the word in the original sentence or the word sequence. The self-attention operation is the weighted dot product of these input vectors $\mathbf{x}_i$ with each other. 

\subsubsection{The Intuition Behind Self-Attention} We can think about the self-attention operation as a two-step process. The first step calculates a normalized dot product between all pairs of input vectors in a given input sequence. The normalization is performed using the softmax operator, which scales a given set of numbers such that the output numbers sum to unity. The normalized correlations are calculated between an input segment $\mathbf{x}_i$ and  all others $j = 1, \ldots, n$: 
\begin{align}
    w_{ij} = \text{softmax}\left(\mathbf{x}_i^T\mathbf{x}_j\right) = \frac{e^{\mathbf{x}_i^T\mathbf{x}_j}}{\sum_k e^{\mathbf{x}_i^T\mathbf{x}_k}},
\end{align}
where $\sum_{j=1}^{n} w_{ij} = 1$ and $1\leq i, j \leq n$. In the second step, for a given input segment $\mathbf{x}_i$, we find a new representation $\mathbf{z}_i$, which is a weighted sum of all input segments $\{\mathbf{x}_i\}_{j=1}^{n}$:

\begin{align}
    \mathbf{z}_i = \sum_{j=1}^{n} w_{ij}\mathbf{x}_j, ~~ \forall ~~ 1\leq i\leq n.
    \label{eq:zi_softmax}
\end{align} 

We note that in Equation \ref{eq:zi_softmax}, for any input segment $\mathbf{x}_i$, the weights $w_{ij}$ add up to 1. Thus, the resulting representation vector $\mathbf{z}_i$ will be similar to the input vector $\mathbf{x}_j$ having the largest attention weight $w_{ij}$. The largest attention weight has, in turn, resulted from the greatest correlation value, as measured by the normalized dot product between $\mathbf{x}_i$ and $\mathbf{x}_j$. Note that $\mathbf{z}_i$ in Equation \ref{eq:zi_softmax} retains the same position in the sequence as $\mathbf{x}_i$. Proceeding further with the next output vector $\mathbf{z}_{i+1}$, the new set of weights corresponding to $\mathbf{x}_{i+1}$ are calculated and used.

\begin{figure}[ht]
  \centering
  \includegraphics[width=1.0\textwidth]{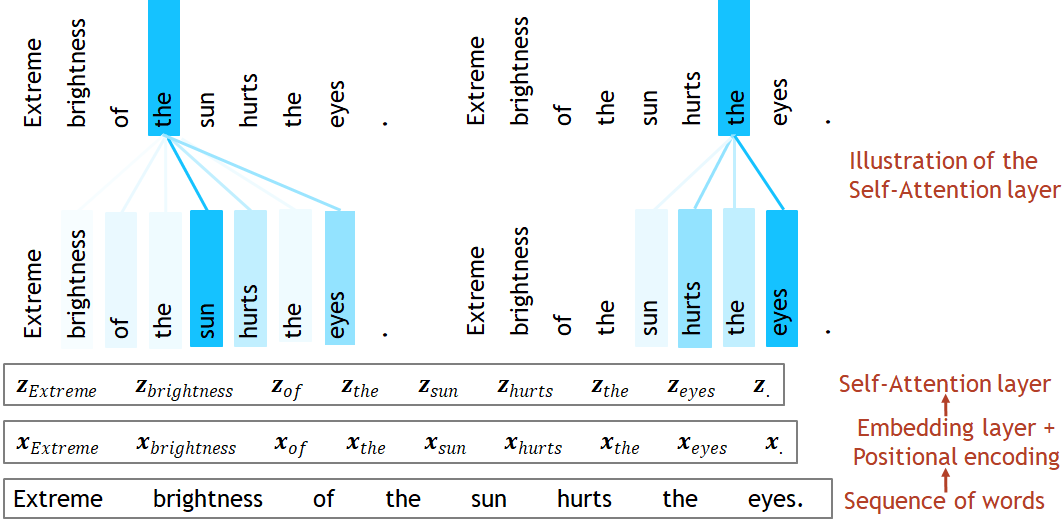}
  \caption{The role of self-attention for the example sentence ``Extreme brightness of the sun hurts the eyes''. The self-attention operation determines the relative correlation of each word with all the words in the sequence. In this example, the first occurrence of the word \emph{the} is most correlated to the word \emph{sun}, whereas the second occurrence of the word \emph{the} has the highest correlation with the word \emph{eyes}. The relative attention weights shown are generated using the Transformer attention visualization tool \cite{bertviz}.}
  \label{fig:self-attention}
\end{figure}

Consider the sentence, ``Extreme brightness of the sun hurts the eyes''. Figure \ref{fig:self-attention} shows the sequence of vector mappings through the self-attention process and provides insight into the role of the self-attention operation. The vector $\mathbf{z}_{sun}$ would be weighted more by the first occurrence of `the' as compared to the second occurrence of `the'. Similarly, $\mathbf{z}_{eyes}$ would be more heavily weighted by $\mathbf{x}_{hurts}$ and $\mathbf{x}_{the}$ (second occurrence of `the') than $\mathbf{x}_{of}$ and the first occurrence of `the'.

\subsubsection{Linearly Weighting Input Using Query, Key, and Value}
The self-attention operation in Transformers starts with building three different linearly-weighted vectors from the input $\{\mathbf{x}_i\}_{i=1}^n$, referred to as query $\mathbf{q} \in \mathbb{R}^{s_1} $, key $\mathbf{k} \in \mathbb{R}^{s_1} $ and value $\mathbf{v} \in \mathbb{R}^s$. Intuitively, a $\mathbf{query}$ is a question that can be one word or a set of words. An example is when one searches for the word `network' on the internet to find more information about it. The search engine maps the query into a set of $\mathbf{keys}$, e.g., neural, social, circuit, deep learning, communication, computer, and protocol. The $\mathbf{values}$ are the candidate websites that have information about the word `network'.

\begin{figure}[ht]
  \centering
  \includegraphics[width=0.5\textwidth]{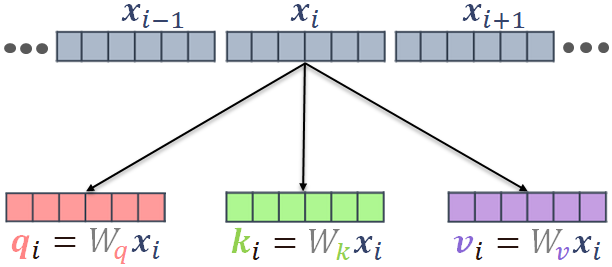}
  \caption{A set of weighted linear transformations applied to the input vector of length $d=6$ are presented. The resulting vectors are referred to as \textbf{query}, \textbf{key}, and \textbf{value}, each of size $s_1=s=d$. There are a total of $n$ such inputs in each sequence processed by the Transformer, which result in $n$ query, $n$ key, and $n$ value vectors.}
  \label{fig:QKV}
\end{figure}

For an input $\mathbf{x}_i$, the query $\color{myorange}\mathbf{q}\normalcolor_i$, key $\color{mygreen}\mathbf{k}\normalcolor_{i}$ and value $\color{mypurple}\mathbf{v}\normalcolor_{i}$ vectors can be found using:
\begin{align}
    \color{myorange}\mathbf{q}\normalcolor_i = W_{\color{myorange}q}\normalcolor\mathbf{x}_i, ~~ \color{mygreen}\mathbf{k}\normalcolor_i = W_{\color{mygreen}k}\normalcolor\mathbf{x}_i, \text{ and } \color{mypurple}\mathbf{v}\normalcolor_i = W_{\color{mypurple}v}\normalcolor\mathbf{x}_i,
\end{align}
where $W_{\color{myorange}q}$ and $W_{\color{mygreen}k} \normalcolor \in \mathbb{R}^{s_1 \times d}$, $W_{\color{mypurple}v} \normalcolor \in \mathbb{R}^{s \times d}$, represent learnable weight matrices. The output vectors $\{\mathbf{z}_{i}\}_{i=1}^n$ are given by,

\begin{equation}
    \mathbf{z}_i=\sum_j \text{softmax}\left(\color{myorange}\mathbf{q}\normalcolor_i^T\color{mygreen}\mathbf{k}\normalcolor_j\right)\color{mypurple}\mathbf{v}\normalcolor_j.
\end{equation}
We note that the weighting of the value vector $\mathbf{v}_{i}$ depends on the mapped correlation between the query vector $\mathbf{q}_i$ at position $i$ and the key vector $\mathbf{k}_j$ at position $j$. The value of the dot product tends to grow with the increasing size of the query and key vectors. As the softmax function is sensitive to large values, the attention weights are scaled by the square root of the size of the query and key vectors $d_q$ as given by
\begin{equation}
    \mathbf{z}_i=\sum_j \text{softmax}\left(\frac{\color{myorange}\mathbf{q}\normalcolor_i^T\color{mygreen}\mathbf{k}\normalcolor_j}{\sqrt{d_q}}\right)\color{mypurple}\mathbf{v}\normalcolor_j.
\end{equation}
In matrix form, we have:
\begin{align} \label{eq:self-attetnion-matrix}
Z = \text{softmax} \left( \frac{QK^T}{\sqrt{d_k}} \right) V,
\end{align}
where $Q$ and $K \in \mathbb{R}^{s_1 \times n}$, and $V \in \mathbb{R}^{s \times n}$, $Z \in \mathbb{R}^{s \times n}$ and $^T$ represents the transpose operation.

\subsection{Multi-Head Self-Attention}
\begin{figure}[ht]
  \centering
  \includegraphics[width=1.0\textwidth]{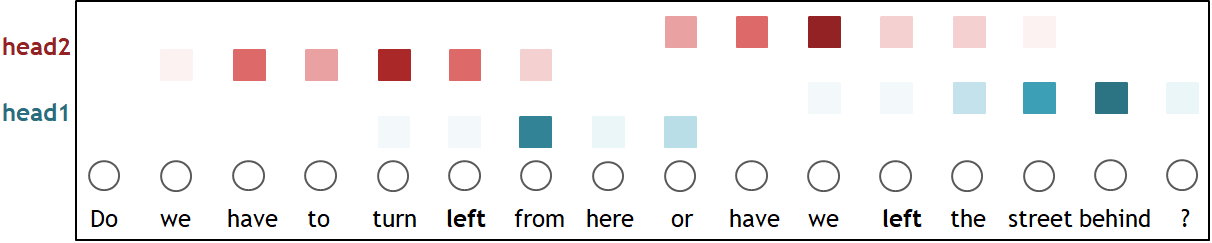}
  \caption{Multi-head self-attention example. Two heads can learn different attention weights for two different uses of the words ``left'' based on their meaning in the sentence. The relative attention weights shown are generated using the Transformer attention visualization tool \cite{bertviz}.}
  \label{fig:multi-head-attention}
\end{figure}

The input data $X$ may contain several levels of correlation information, and the learning process may benefit from processing the input data in different ways. Multiple self-attention heads are introduced that operate on the same input in parallel and use distinct weight matrices $W_q, W_k,$ and  $W_v$ to extract various levels of correlation between the input data. For example, consider the sentence ``Do we have to turn left from here or have we left the street behind?''. There are two occurrences of the word ``left'' in the sentence. Each occurrence has a different meaning and, consequently, a different relationship with the rest of the words in the sentence. As shown in Figure \ref{fig:multi-head-attention}, Transformers can capture such information using multiple heads. Each head is built using a separate set of query, key, and value weight matrices and computes self-attention over the input sequence in parallel with other heads. Using multiple heads in a Transformer is analogous to using multiple kernels at each layer in a convolutional neural network, where each kernel is responsible for learning distinct features or representations \cite{albawi2017understanding}.

\begin{figure}[ht]
  \centering
  \includegraphics[width=0.97\textwidth]{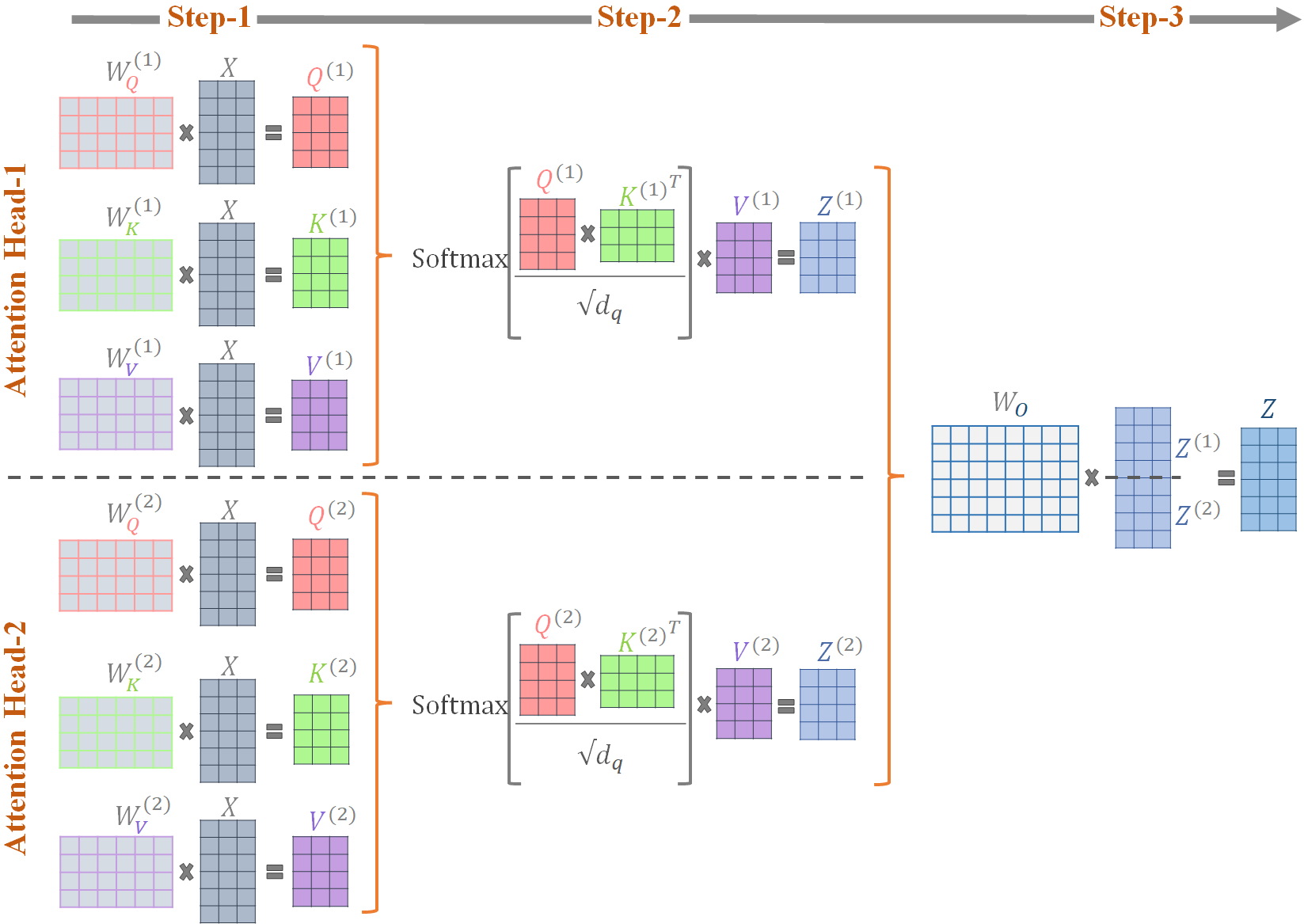}
  \caption{Multi-head attention steps. The example consists of two heads, and the input sequence comprises of three words. The input $\{\mathbf{x}_i\}_{i=1}^n$ is mapped to word embedding, and positional encoding vectors are added, resulting in the input matrix $X \in \mathbb{R}^{6 \times 3}$. Two sets of query, key, and value matrices, $Q^{(l)}, K^{(l)},$ and $V^{(l)} \in \mathbb{R}^{4 \times 3}$ ($l=2$), are constructed corresponding to each attention head. The self-attention operation applied on both sets of query, key, and value matrices produces output matrices $Z^{(1)}$ and $Z^{(2)} \in \mathbb{R}^{4 \times 3}$. A linear operation, involving a learnable parameter $W_o \in \mathbb{R}^{d \times rs}$, follows the concatenation of $Z^{(1)}$ and $Z^{(2)}$ to map it to the same dimension as input matrix $X$. Output matrix $Z \in \mathbb{R}^{6 \times 3}$ accumulates information from all the attention heads.}
  \label{fig:Multihead_step1}
\end{figure}

The following three steps can describe the operations involved in multi-head self-attention.

\subsubsection{Step 1 - Generation of Multiple Sets of Distinct Query, Key, and Value Vectors}
Assuming we have a total of $r$ heads,  a total of $r$ sets of weight matrices $\{W_q^{(l)}, W_k^{(l)}, W_v^{(l)}\}_{l=1}^r$ will generate $r$ sets of distinct query, key and value matrices for the input $X$. The process is illustrated in Figure \ref{fig:Multihead_step1} for the case of an input with three vectors ($n=3$) of dimension six each ($d=6$) and $s_1=s=4$. This leads to the input matrix $X \in \mathbb{R}^{6 \times 3}$, $W_q^{(l)}, W_k^{(l)}, W_v^{(l)} \in \mathbb{R}^{4 \times 6}$ and $Q^{(l)}, K^{(l)},$ and $V^{(l)} \in \mathbb{R}^{4 \times 3}$.

\subsubsection{Step-2 - Scaled Dot Product Operations in Parallel} This step consists of implementing the following relationship as shown in Figure \ref{fig:Multihead_step1}: 
\begin{align} \label{eq:self-attetnion-matrix11}
Z = \text{softmax} \left( \frac{QK^T}{\sqrt{d_k}} \right) V.
\end{align}

\subsubsection{Step-3 - Concatenating and Linearly Combining Outputs} Finally, we concatenate outputs $Z^{(l)}$ from all $r$ heads and linearly combine using a learnable weight matrix $W_o \in \mathbb{R}^{d \times rs}$. The output is a matrix $Z \in \mathbb{R}^{d \times n}$. It is important to note that the input and output of the multi-head self-attention are of the same dimension, that is, dimension $(X) =$ dimension $(Z)$.

\begin{figure}[ht]
  \centering
  \includegraphics[width=1\textwidth]{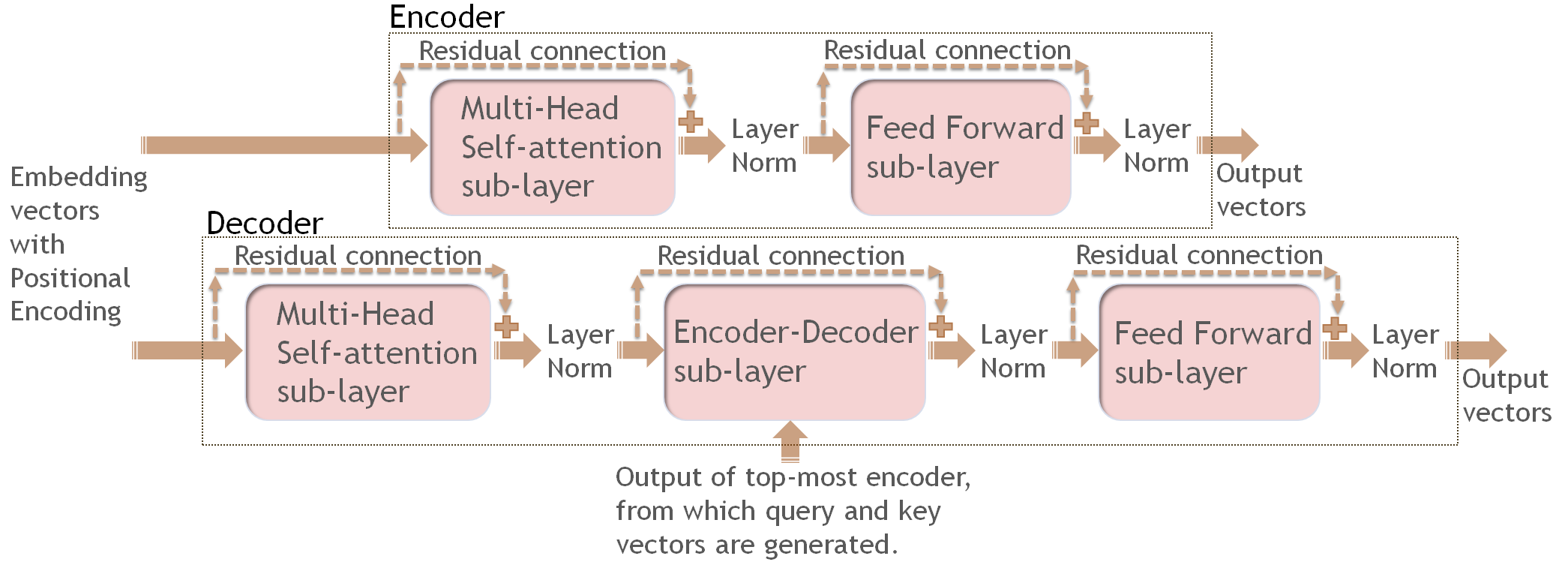}
  \caption{An encoder and decoder block is presented. Each encoder block is built using multi-head self-attention and feed-forward layers with residual connections around each layer followed by the layer normalization operation. The decoder block is similar to the encoder block with an additional encoder-decoder attention layer. This encoder-decoder attention layer receives the output of the topmost encoder block and uses it to generate the key and value vectors. The query vectors are produced from the output of the multi-head self-attention layer preceding the encoder-decoder layer.}
  \label{fig:Transformer_block}
\end{figure}

\subsection{Building Transformers Using Encoders and Decoders}
\label{subsubsection:encoder}
The Transformer architecture is generally composed of multiple instances of two types of components referred to as encoders and decoders. 

\subsubsection{The Encoder Block}
As shown in Figure \ref{fig:Transformer_block}, an encoder block consists of a multi-head self-attention layer and a feed-forward layer connected back-to-back with residual connections and normalization layers. Residual connections are a commonly used technique for training deep neural networks and help in training stabilization and learning \cite{szegedy2015going}. The layer normalization operation is also commonly used in neural networks for processing sequential data. It helps with the faster convergence of the model training \cite{ba2016layer}. The feed-forward layer comprises two linear layers with a ReLU activation function \cite{agarap2018deep}. The output of an encoder block is used as an input to the next encoder block. The input to the first encoder block consists of the sum of word embeddings and positional encoding (PE) vectors.

\subsubsection{The Decoder Block}
Each decoder block consists of similar layers and operations as the encoder block. However, a decoder receives two inputs, one from the previous decoder and the second from the last encoder. Inside a decoder, three layers include (1) multi-head self-attention, (2) an encoder-decoder attention layer, and (3) a feed-forward layer. There are residual connections and layer normalization operations, as shown in Figure \ref{fig:Transformer_block}. Inside the encoder-decoder attention layer, a set of key and value vectors are generated from the output of the last encoder. The query vectors are produced from the output of the multi-head self-attention layer preceding the encoder-decoder layer.

\subsubsection{Masking in Self-Attention}
Inside the decoder block, the multi-head self-attention layer masks parts of the target input during the training phase. This operation ensures that the self-attention operations do not involve future data points, i.e., the values the decoder is expected to predict. During the training phase, the model's predicted output is not fed back into the decoder. Instead, the ground truth target (word embedding) is used to aid the learning. During the testing phase, the predicted words in the sequence are fed back to the decoder after passing them through a word embedding layer and the addition of PE, as shown in Figure \ref{fig:encoder-decoder-stack}.

\subsubsection{Stacking Encoders and Decoders}
\begin{figure}[ht]
  \centering
  \includegraphics[width=0.97 \textwidth]{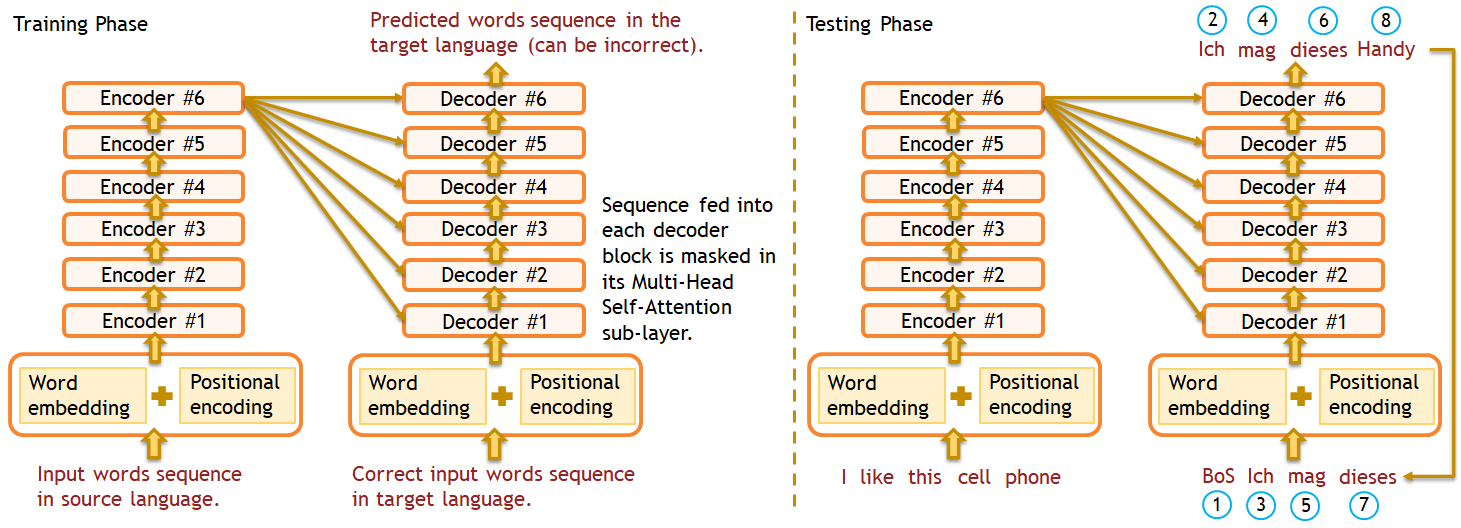}
  \caption{Stacked encoder and decoder blocks used in Transformers are presented for the machine translation task. The number of stacked blocks may vary depending on the task, analogous to multiple layers used in traditional neural networks. Input to the stack of encoders is word embeddings enriched with positional information. The output of the topmost encoder is fed into each decoder, which helps develop the relationship between the source and target language. The output of the final decoder, mapped to the desired target language vocabulary, predicts the next word in the sequence. During training, unlike the testing phase, the correct sequence is fed back into the decoder stack irrespective of the predicted word.} 
  \label{fig:encoder-decoder-stack}
\end{figure}

A Transformer model may contain stacks of multiple encoder and decoder blocks depending on the problem being solved, as shown in Figure \ref{fig:encoder-decoder-stack} \cite{tschannen2018recent}. The stacked encoder/decoder blocks resemble multiple hidden layers used in traditional neural networks. However, it is important to note that, generally, there is no reduction in the representation dimensions after processing by the encoder or the decoder. The input to the first encoder block is the word sequence mapped into word embeddings with PEs added.

\subsubsection{The Output}
The output from the last encoder is fed into each decoder along with the input from the previous decoder. Optionally, the output of the last decoder block is passed through a linear layer to match the output dimension to the desired size, e.g., target language vocabulary size. The mapped output vectors are then passed through a softmax layer to find the probability of the next word in the output sequence. Depending on the desired task, the operations on the output from the last decoder block can be selected for classification or regression.

\subsection{Positional Encoding (PE)}
\label{section:PE}
The most important aspect of processing sequential data is incorporating the order of the sequence. The self-attention operations do not include any information about the order of the input data in a sequence. Transformers (1) use the concept of positional encoding to add the positional information to the input and (2) process the modified input in parallel, thereby avoiding the challenges of processing data sequentially. The technique consists of calculating $n$ PE vectors (denoted as $p \in \mathbb{R}^{d} $) and adding these to the input $\{\mathbf{x}_i\}_{i=1}^n$.  

\subsubsection{Sinusoidal PE}
\begin{figure}[ht]
  \centering
  \includegraphics[width=1.0 \textwidth]{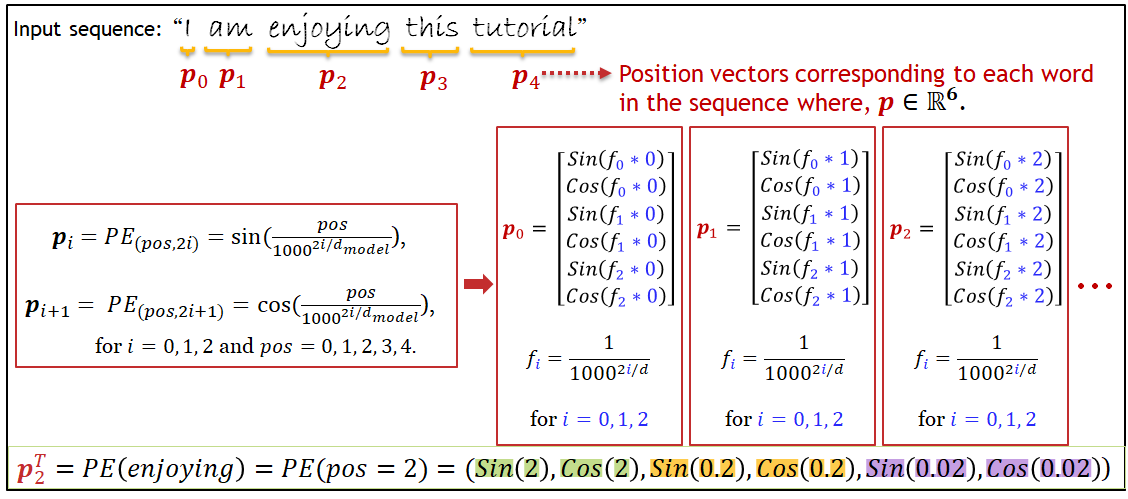}
  \caption{A set of positional encoding vectors are presented for an example input sequence. In this example, the formulation of PE vectors uses a value of $1,000$ instead of $10,000$ (as in the original Transformer implementation).}
  \label{fig:pemechanics}
\end{figure}

In the original work, the authors proposed sinusoidal functions for pre-calculating PE vectors for the input dataset \cite{Vaswani2017}. The PE vectors do not include any learnable parameters and are added directly to the word embedding input vectors. Figure \ref{fig:pemechanics} shows the formulation of PE vectors with Equations \ref{eq:PE-1} and \ref{eq:PE-2},

\begin{alignat}{2}
    \text{PE}_{(pos, 2i)} &= \mathbf{p}_i &&= \sin\left(\frac{pos}{10000^\frac{2i}{d}}\right), \label{eq:PE-1} \\
    \text{PE}_{(pos, 2i+1)} &= \mathbf{p}_{i+1}  &&= \cos\left (\frac{pos}{10000^\frac{2i}{d}}\right), \label{eq:PE-2}
\end{alignat}

where $pos$ is the position (time-step) in the sequence of input words, $i$ is the position along the embedding vector dimension ranging from 0 to $\frac{d}{2}-1$, and $d$ represents the dimension of the embedding vectors. The number 10,000 used in Equations \ref{eq:PE-1} and \ref{eq:PE-2} can differ depending on the input sequence's length.

\subsubsection{Relationship of the Sinusoidal PE with Binary Encoding} 
\begin{figure}[ht]
  \centering
  \includegraphics[width=1 \textwidth]{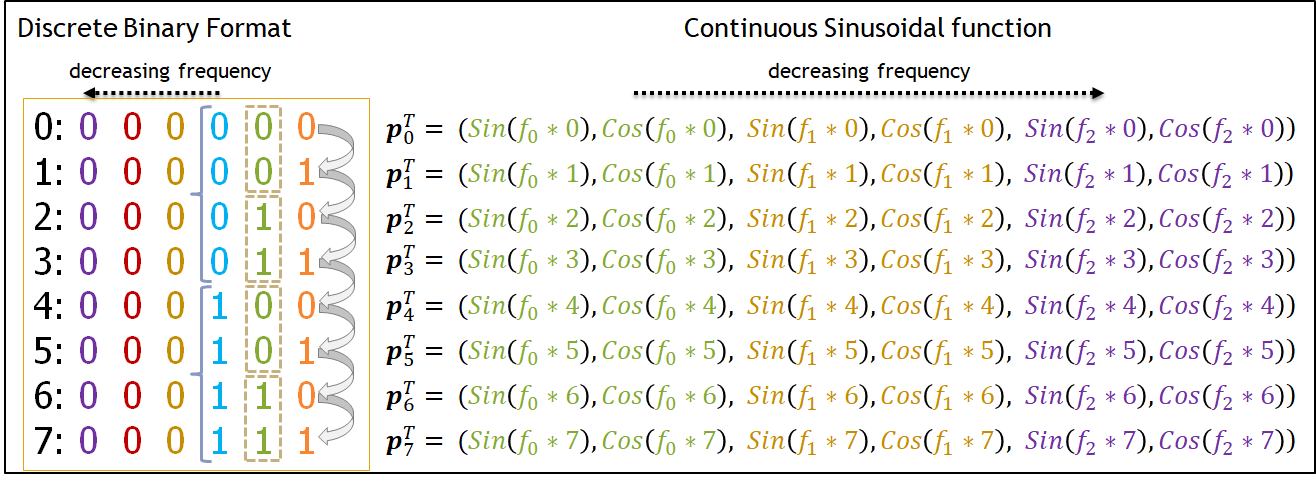}
  \caption{An analogy between binary format and sinusoidal function for positional encoding is presented. In the binary format, the frequency of alternating bits decreases as we move from orange to purple bits. Sinusoidal PE shows a similar behavior of changing frequency as we move along the positional encoding vector. We consider an input sequence of length $n=8$ and a word embedding vector of size $d=6$.}
  \label{fig:analogype}
\end{figure}

We can draw an analogy between the proposed sinusoidal functions for PE and alternating bits in a six-bit long binary number, as shown in Figure \ref{fig:analogype} \cite{Vaswani2017,takase2019positional}. In the binary format, the least significant bit (shown in the orange color) alternates with the highest frequency. Moving to the left (indigo), the frequency of bit oscillation between $0$ and $1$ decreases. Similarly, in the sinusoidal PE, as we move along the positional encoding vector, the frequency of the sinusoidal function changes.

\subsubsection{Positional Encoding and Rotation Matrix}
The PE vectors calculated using sinusoidal functions allow the model to learn about the relative position of words rather than their absolute position. For example, in the sentence ``I am enjoying this tutorial'', the absolute position of the word `this' is $4$. However, relative to the word `am', the word `this' is at position $3$. Hence, for any fixed offset $k$, the PE vector $\mathbf{p}_{i+k}$ can be represented as a linear transformation of $\mathbf{p}_i$. Consider $\mathbf{p}_i \in \mathbb{R}^2$ and let $T=\begin{pmatrix} x_1 & y_1 \\ x_2 & y_2 \end{pmatrix} $ be a linear transformation with $a$ representing the absolute position of a word in an input sequence. We can write 

\begin{align}
    T\begin{bmatrix}
        \sin\left(f_ia\right)\\
        \cos\left(f_ia\right)
     \end{bmatrix} & = 
     \begin{bmatrix}
        \sin\left(f_i\left(a + k\right)\right)\\
        \cos\left(f_i\left(a + k\right)\right)
    \end{bmatrix}, \\
    \begin{bmatrix}
        x_1 & y_1\\
        x_2 & y_2
    \end{bmatrix}
    \begin{bmatrix}
        \sin(f_ia)\\
        \cos(f_ia)
    \end{bmatrix} &= 
    \begin{bmatrix}
        \sin(f_i(a + k))\\
        \cos(f_i(a + k))
    \end{bmatrix},\\
    \begin{bmatrix}
        x_1 \sin(f_ia) + y_1 \cos(f_ia)\\
        x_2 \sin(f_ia) + y_2 \cos(f_ia)
    \end{bmatrix} &=
    \begin{bmatrix}
        \sin(f_ia) \cos(f_ik) + \sin(f_ik) \cos(f_ia)\\
        \cos(f_ia) \cos(f_ik) - \sin(f_ia) \sin(f_ik)
    \end{bmatrix}. \label{eq:PE-transformation}
\end{align}
Comparing both sides of Equation \ref{eq:PE-transformation}, we get $x_1 = \cos(f_ik)$, $y_1= \sin(f_ik)$, $x_2=- \sin(f_ik)$, and $y_2=\cos(f_ik)$. The transformation $T$ can now be written as:
\begin{equation}
    T=\begin{bmatrix}
        \cos(f_ik) & \sin(f_ik)\\
        -\sin(f_ik) & \cos(f_ik)
        \end{bmatrix}.
\end{equation}
We note that the transformation $T$ is a rotation matrix and depends on the relative position of words, not the absolute position. The sinusoidal PE function works for an input sequence of any length that need not be specified.

\subsubsection{Combining Positional Encoding with Word Embeddings}
The PE vectors are added to the word embeddings for each word in the input sequence. We may argue that the addition operation may result in the loss of some information from both sources, i.e., PE and word embeddings. However, that may not be the case, given that both PE and word embeddings encode different types of information. The PE vectors contain information about the location of a word in the input sequence, while the word embeddings encode semantic and contextual information about the word. The two encoding schemes belong to different sub-spaces, which may be orthogonal to each other. In this case, the addition of two such vectors may not result in the loss of information. We can consider an analogy with the frequency modulation operation as done in digital communication --- a signal of lower frequency rides over a high-frequency signal without the two interfering with each other.

\begin{figure}[htb]
     \centering
     \centering
     \includegraphics[width = 0.98\linewidth]{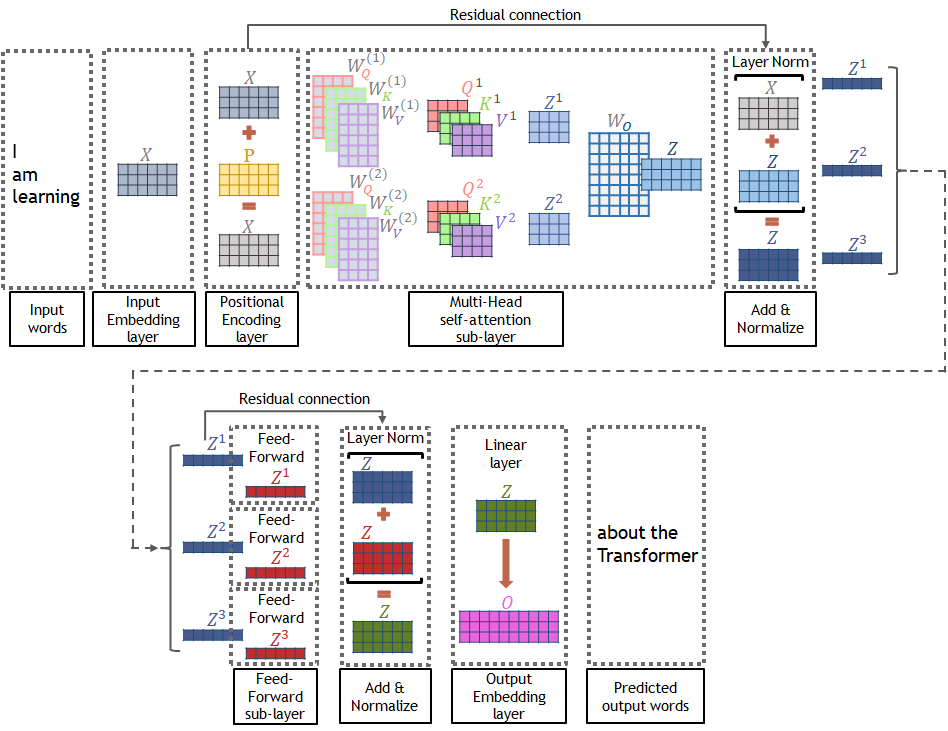}
     \caption{An example of a Transformer model with a single encoder is presented. The end-to-end sequence of operations is shown for an input sequence of three words and an output prediction of the expected three following words.}
     \label{fig:transformer}
 \end{figure}

Now that we have discussed each operation individually as implemented in the Transformer architecture, Figure \ref{fig:transformer} depicts the end-to-end flow of the internal operations in the Transformer architecture with a single encoder and three input words to predict the next three words.
\newpage
 \newpage
\section{Road Map of Transformers for Time-Series Analysis} \label{section:categories}
Since the inception of Transformers in 2017 \cite{Vaswani2017}, there have been many advances in these networks for time-series analysis. The original model mainly focuses on NLP tasks, but now the architecture has been expanded for classification \cite{yuan2021tokens}, time-series analysis \cite{song2017attend}, semantic segmentation \cite{zheng2021rethinking}, and more. Time-series data were not part of the initial conception of Transformers. Many researchers have customized and improved the architecture’s performance for time-series analysis. In order to illustrate how this research has unfolded, we will provide a road map of time-series tasks and how the technology has advanced. 

A commonality amongst the improvements is that the input layer is modified to accommodate time-series data. The use of Transformers for time-series analysis, forecasting, and classification accomplishes two main tasks. Within each task, we provide useful information and links to the datasets used in the state-of-the-art methods. The road map in this section provides a comprehensive breakdown of the advancements made in the past few years and their relation to each other. Towards the end of each subsection is a list of the models (and citations) which are included in that category.

\subsection{Avenues of Improvement for Time-Series Transformers} \label{section:avenues}
\begin{figure}[ht]
  \centering
  \includegraphics[width=0.95\textwidth]{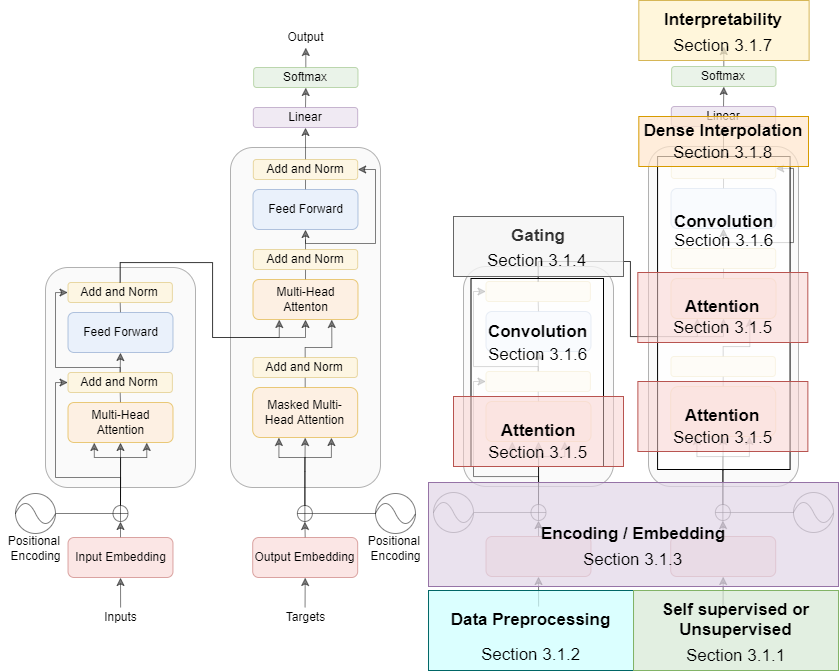}
  \caption{Overview of avenues of improvement for time-series Transformers. On the left is a recreation of the original Transformer architecture \cite{Vaswani2017}. The right side shows locations in the original architecture for each of the avenues of improvement in subsection \ref{section:avenues} with respect to the original Transformer model.}
  \label{fig:categories_section_3}
\end{figure}

Each part of this subsection outlines major contributions and research articles that have improved specific mechanisms within the Transformer architecture. A schematic road map for studying Transformers in time-series analysis is shown in Figure \ref{fig:categories_section_3}. 

\subsubsection{Learning Type - Supervised, Self-supervised, or Unsupervised}
Much of the mainstream Transformer applications rely on the training data, which is hand labeled. Labeling data can take a considerable time, rendering the massive amount of unlabeled data unusable. Self-supervised and unsupervised methods seek to improve Transformers by allowing them to learn, categorize, and forecast data that has not been labeled. For example, there is a considerable amount of unlabeled satellite imaging data. In \cite{yuan2020self}, a proposed  model, SITS-BERT, learns from unlabeled data to apply region classification from satellite imagery. Other self-supervised or unsupervised learning models include anomaly Transformer \cite{xu2021anomaly} and self-supervised Transformer for Time-Series (STraTS) \cite{tipirneni2021self}.

\subsubsection{Data Preprocessing}
Preprocessing is routinely performed to prepare the data to be fed into a machine learning model. Preprocessing operations are known to impact the performance of machine learning models. Some researchers either add noise to input data features \cite{yuan2020self}, perform masking \cite{shankaranarayana2021attention}, or variable selection \cite{lim2019temporal}. The operation of masking removes features in the input thereby improving performance by making the model better at predicting missing features. A similar concept applies when adding noise to the input for training, except that the features become relatively noisier but are not replaced entirely. Both these operations can improve the robustness of the model so that it attains good accuracy despite the noise in the training or test data.

\subsubsection{Positional Encoding (PEs)}
Some recent work has focused on improving upon the original PEs proposed in\cite{Vaswani2017}. Transformers use PEs to embed sequence/time information in the model input so that Transformers can process sequential data all at once, rather than one at a time like an RNN, LSTM, or GRU. Embedding time (seconds, minutes, hours, weeks, years, etc.) into the input allows the model to better analyze time-series data and leverage computational benefits offered by modern hardware, including GPUs, TPUs, and others. Recently, timestamp embedding \cite{shankaranarayana2021attention}, temporal encoding \cite{cai2020traffic}, and other methods have been proposed for creating Transformers which can be trained more efficiently \cite{lim2019temporal}.

\subsubsection{Gating Operation}
A gating operation merges the output of two encoder towers of a standard Transformer model for a single set of output predictions \cite{liu2021gated}. This operation combines and selects multiple outputs from an encoder or decoder block. Gating also benefits from applying non-linear data processing when appropriate. Application of gating in various ways is a likely avenue for future innovations in Transformers, not only for time-series data but also for any other type of data. Several architectures use gating, including Gated Transformer Networks (GTN) \cite{liu2021gated} and Temporal Fusion Transformers (TFT) \cite{lim2019temporal}. The proposed architecture of GTN uses gating techniques to incorporate information from two Transformer towers into the output. It does this by concatenating the output of each tower. The TFT proposes Gated Linear Units (GLUs) \cite{dauphin2017language} to allow different parts of the network to be emphasized or suppressed based on the dataset.

\subsubsection{Attention}

The models discussed in this section improve upon Transformers for time-series data by modifying and improving the attention mechanisms of the model.

The Tightly-Coupled Convolutional Transformer (TCCT) \cite{shen2022tcct} proposes three architectures that improve upon attention. The first is called Cross Stage Partial Attention (CSPAttention). This approach combines Cross Stage Partial Network (CSPNet) \cite{Wang_2020_CVPR_Workshops} with a self-attention mechanism to reduce required resources. CSPNet reduces computations by incorporating the feature map at the input into the output stage. CSPAttention applies this concept to just the attention layers, considerably reducing the time complexity and memory required. The second approach changes the self-attention distilling operation and how the self-attention blocks connect. Instead of canonical convolution, they use dilated causal convolution. Dilated causal convolution enhances the locality and allows the Transformer to attain exponentially receptive field growth. The third architecture is a pass-through mechanism that allows multiple self-attention blocks to be stacked. This stacking is done by concatenating feature maps from multiple scales within the self-attention mechanisms. This mechanism improves the ability of the Transformer to pick up on fine-scale features.

More approaches to improving attention include Non-Autoregressive Spatial-Temporal Transformer (NAST) \cite{chen2021nast} and Informer \cite{haoyietal-informer-2021}. The architecture proposed in NAST is referred to as the Spatial-Temporal Attention Block. This approach combines a spatial attention block (which makes predictions across space) with a temporal attention block (which makes predictions across time). The result improves learning in both the spatial and temporal domains. The Informer architecture replaces the canonical self-attention with the ProbSparse self-attention mechanism, which is more efficient in time complexity and memory usage. This Transformer also uses a self-attention distilling function to decrease space complexity. These two mechanisms make the Informer highly efficient at processing exceedingly large input data sequences.

Lastly, we will discuss three architectures that improve upon attention, LogSparse Transformers \cite{li2019enhancing}, TFT \cite{lim2019temporal}, and YFormer \cite{madhusudhanan2021yformer}. LogSparse Transformers introduce convolutional self-attention blocks, consisting of a convolutional layer prior to the attention mechanism for creating the queries and keys. A temporal self-attention decoder is used in TFT for learning the long-term dependencies in the data. The YFormer architecture proposes a sparse attention mechanism combined with a downsampling decoder. This architecture allows the model to better detect long-range effects in the data.

\subsubsection{Convolution}
The original Transformer architecture does not make use of convolutional layers. However, this does not mean that Transformers would not benefit from the addition of convolutional layers. In fact, many Transformers designed for time-series data have benefited from adding convolutional layers or incorporating convolution into existing mechanisms. Most approaches incorporate convolutions either prior to or alongside the attention mechanism within the Transformer.

Approaches that improve upon Transformers via convolution include TCCT \cite{shen2022tcct}, LogSparse Transformers \cite{li2019enhancing}, TabAConvBERT \cite{shankaranarayana2021attention}, and Traffic Transformers \cite{cai2020traffic}. The TCCT uses a mechanism called dilated causal convolution from the Informer \cite{haoyietal-informer-2021} architecture, replacing the canonical convolutional layers. LogSparse Transformers also use causal convolutional layers. As mentioned in the previous section, this layer generates queries and keys for the self-attention layers, called convolutional self-attention. TabAConvBERT employs one-dimensional convolutions considering that it is naturally effective for time-series data. Traffic Transformers \cite{cai2020traffic} incorporate concepts from graph neural networks into the convolutional layers to produce Graph Convolutional Filters.

\subsubsection{Interpretability/Explainability}
\label{subsection:Interpretability}
Transformers are a relatively new class of machine learning models compared to CNNs or LSTMs. For their reliable and trustworthy use, we must understand the black-box nature of these models and explain their decisions.
The black-box phenomenon is a prevalent problem in artificial intelligence. This refers to the situation where only the inputs and outputs of a learning model can be observed. It is not precisely known how the parameters of the model interact to arrive at the final output.

Many approaches to interpreting and explaining model predictions are post hoc, that is, the explanation is made after the fact. Post hoc approaches apply to almost any model. Many of these approaches provide visually appealing results, but might not accurately explain what is occurring inside the model \cite{nielsen2021robust}. One possible approach is to incorporate explanations and interpretability into the model itself, rather than being approximated after the fact. There now exist multiple time-series Transformers that are inherently interpretable \cite{tipirneni2021self, liu2021gated, lim2019temporal}. These models can produce explanations that allow for better interpretations of results and greater user trust.

\subsubsection{Dense Interpolation}
The Transformer model generally consists of encoder and decoder blocks followed by linear and softmax layers for decision-making. One approach referred to as Simply Attend and Diagnose (SAnD) replaces the decoder block with a dense interpolation layer to incorporate temporal order into the model’s processing \cite{song2017attend}. This approach does not utilize output embedding, thereby cutting down the number of total layers in the model. Without the decoder, the model needs a way to process the outputs of the encoder block to be input into the linear layers. Simple concatenation leads to poor prediction accuracy. Therefore, the work in \cite{song2017attend} developed a dense interpolation algorithm with hyperparameters that can be tuned to improve performance.

\subsection{Architectural Modifications}
\subsubsection{BERT-Inspired}
A famous architecture that builds on the original Transformer paper is Bidirectional Encoder Representations from Transformers (BERT) \cite{Devlin2019BERT:Understanding}. The model is built by stacking the Transformer encoder blocks and introducing a new training scheme. The encoder block is pre-trained independently of the task. The decoder block can be added later and fine-tuned for the task at hand. This scheme allows training BERT models on large amounts of unlabeled data.

The BERT architecture has inspired many new Transformer models for time-series data \cite{yuan2020self, shankaranarayana2021attention, qi2021known, xu2021anomaly}. Creating a BERT-style model for time-series data has some challenges compared to NLP tasks. Language data is a standardized type of data that can be used for various tasks, including translation, text summarization, question answering, sentiment analysis, etc. All of these tasks can use the same data for pre-training. However, this is not the case for the time-series tasks. Examples of time-series data include electricity usage \cite{Dua2019}, ambient temperature \cite{haoyietal-informer-2021}, traffic volume \cite{Dua2019}, satellite imagery \cite{yuan2020self}, various forms of healthcare data \cite{johnson2016mimic}, and more. With this variety of data types, the pre-training process will have to be different for each task. This task-dependent pre-training contrasts with the NLP tasks which can start with the same pre-trained models assuming all tasks are based on the same language semantics and structure.

\subsubsection{GAN-Inspired}
Generative adversarial networks (GANs) consist of two deep neural networks, the generator, and the discriminator. Both networks learn adversarially from each other. GANs are commonly used in image processing for generating realistic images. The generator's task is to create images that will trick the discriminator. The discriminator is given the actual and generated (fake) images and must predict whether the input image was real or fake. A well-trained GAN can generate images that look very realistic to a human. 

The same generator-discriminator learning principle has been applied to the time-series forecasting task \cite{wu2020adversarial}. The authors use a Transformer as the generator and a discriminator and train the model for accurate forecasting predictions. The generator's task is to create a forecast that the discriminator will classify as real or fake. As the training continues, the generator network will create more realistic data. At the end of the training, the model will be highly accurate at making predictions.

\subsection{Time-Series Tasks}
The two main tasks performed on time-series data are forecasting and classification. Forecasting seeks to predict real-valued numbers from given time-series data, referred to as regression. Many forecasting Transformers for time-series data have been developed in the recent literature \cite{shen2022tcct, chen2021nast, haoyietal-informer-2021, li2019enhancing, wu2020deep, lim2019temporal, cai2020traffic, wu2020adversarial, qi2021known, tipirneni2021self, madhusudhanan2021yformer}. The classification task involves categorizing the given time-series data into one or more target classes. There have been many recent advancements for time-series Transformers for classification tasks \cite{shankaranarayana2021attention, xu2021anomaly, yuan2020self, song2017attend, liu2021gated}. All of the time-series Transformer-based models discussed in this tutorial focus on one of these two tasks. Some of these models can accomplish both tasks after small modifications in the last layer and the loss function.

\section{Time-Series Analysis - Architectures and Use Cases} \label{section:time-series}

\subsection{The Informer Architecture}
Recently, Zhou et al. proposed \emph{Informer}, which uses a \textit{ProbSparse} self-attention mechanism to optimize the computational complexity and memory usage of the standard Transformer architecture \cite{haoyietal-informer-2021}. The authors also introduced the self-attention distilling operation, which considerably reduces the total space complexity of the model.

\begin{figure}[ht]
  \centering
  \includegraphics[width=0.9\textwidth]{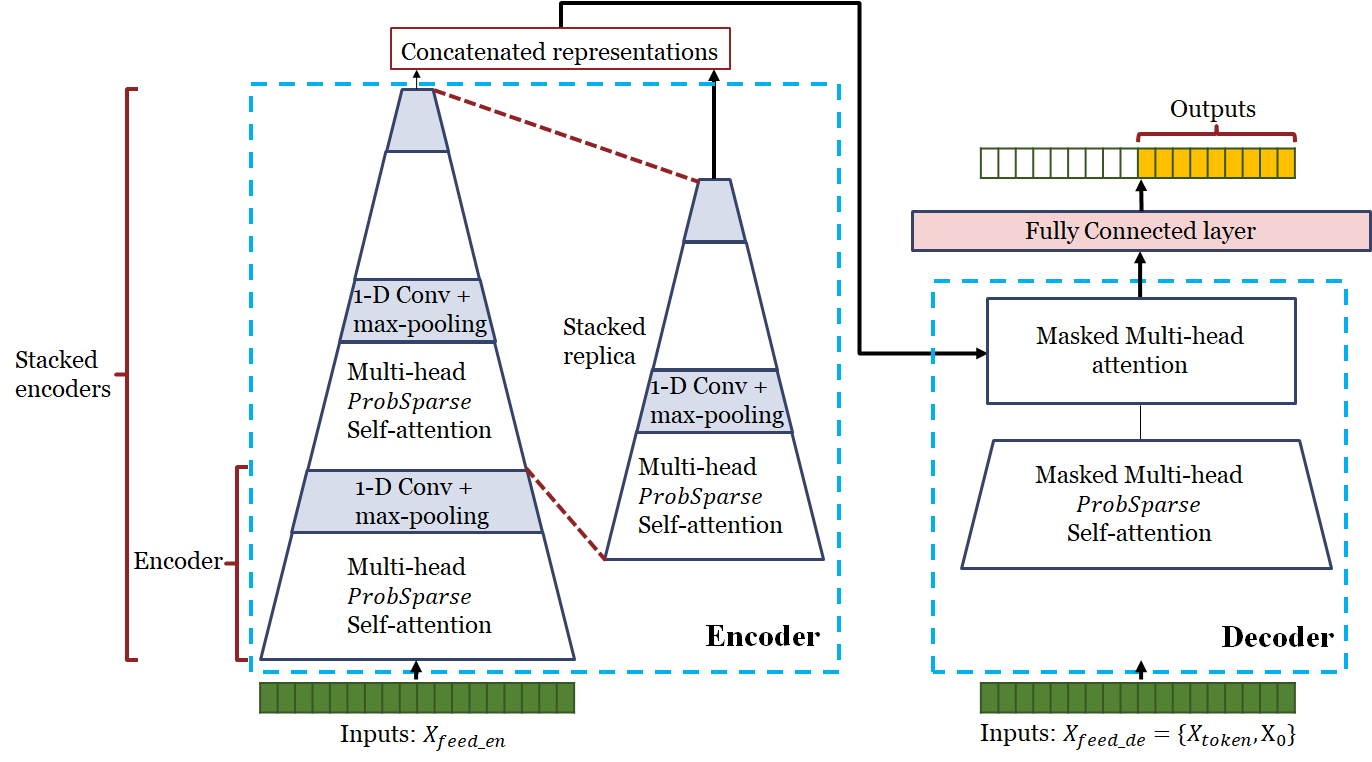}
  \caption{Informer architecture as proposed in \cite{haoyietal-informer-2021} is presented. Informer consists of stacked encoders and decoders. The encoder comprises multi-head ProbSparse self-attention and self-attention distilling operations. Stacked replicas are built with an input length of half of the previous layer. The encoder output is a concatenation of the outputs from the main and replica stacks. The decoder has two stacked multi-head attention layers and predicts the entire output sequence simultaneously in one forward pass.}
  \label{fig:Informer_architecture}
\end{figure}

ProbSparse self-attention uses the dominant dot-product pairs by randomly selecting $\log L$ top query vectors and setting the rest of the query vector values to zero. The computational complexity and memory usage reduce with $L$ value vectors to $\mathcal{O}(L\log L)$. With a stack of $J$ encoders, the total memory usage reduces to $\mathcal{O}(J L\log L)$. Furthermore, the self-attention distilling operation removes redundant combinations of value vectors. This operation is inspired by dilated convolution as proposed in \cite{8099558} and \cite{gupta2017dilated}. The output of the multi-head self-attention is fed into 1-D convolution filters with kernel size equal to 3. Later, an exponential linear unit (ELU) activation function is applied, followed by a max-pooling operation with a stride of 2. These operations reduce the size in half and thus, form a pyramid as shown in Figure \ref{fig:Informer_architecture}. Effectively, the total space complexity reduces considerably. Stacked replicas are also built, with an input length of half of the previous stack. Figure \ref{fig:Informer_architecture} shows only one replica stack. The output of the main stack and the replica stacks have the same dimension and are concatenated to form the final output of the encoder. These replica stacks enhance the robustness of the distilling operation.

The decoder consists of two stacked multi-head attention layers. The input to the decoder is a start token concatenated with a placeholder for the predicted target sequence (with initial values set to zero). It predicts all outputs by one forward procedure (as shown in Figure \ref{fig:Informer_architecture}) thereby considerably reducing inference time.

In the Informer architecture, the scalar input is mapped to $d$ dimensional vectors $\mathbf{u}_i$, using 1-D convolution filters. The local context is retained using the fixed positional encoding (PE) based on sinusoidal functions. A global timestamp, called stamp embedding (SE), is also included to capture hierarchical time information such as the week, month, or year as well as occasional events such as holidays. The input to the encoder is a sum of the scalar projection vectors ($\mathbf{u}_i$), $\text{PE}_i$, and SE.

The Informer architecture was tested on various datasets, including electricity consumption load and the weather dataset. The model performed better than state-of-the-art (SOTA), including Autoregressive Integrated Moving Average (ARIMA) \cite{ariyo2014stock}, Prophet \cite{taylor2018forecasting}, LSTMa \cite{bahdanau2014neural}, LSTnet \cite{lai2018modeling}, and DeepAR \cite{salinas2020deepar}. 

\subsection{LogSparse Transformer Architecture}
\begin{figure}[ht]
  \centering
  \includegraphics[width=1.0\textwidth]{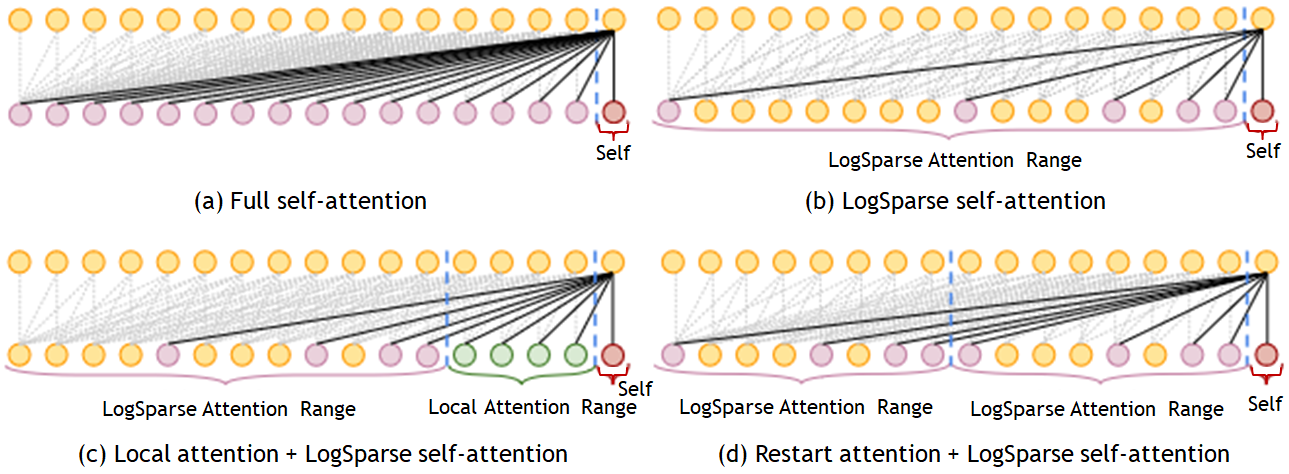}
  \caption{Various ways to apply LogSparse self-attention as proposed by Li et al.\cite{li2019enhancing}.}
  \label{fig:Logsparse}
\end{figure}

Li et al. proposed LogSparse Transformers to overcome memory challenges, thus making Transformers more feasible for time-series data with long-term dependencies \cite{li2019enhancing}. LogSparse Transformers allow each time step to attend to previous time steps that are selected using an exponential step size. This reduces the memory utilization from $\mathcal{O}(L^2)$ to $\mathcal{O}(L\log_2 L)$ in each self-attention layer. Figure \ref{fig:Logsparse} shows the various ways in which LogSparse self-attention can be used for time-series analysis. The canonical self-attention mechanism used for neighboring time steps in a specific range allows for gathering more information. Beyond that range, the LogSparse self-attention mechanism is applied. Another way is to restart the LogSparse step size after a particular range of time steps.

\begin{figure}[ht]
  \centering
  \includegraphics[width=1.0\textwidth]{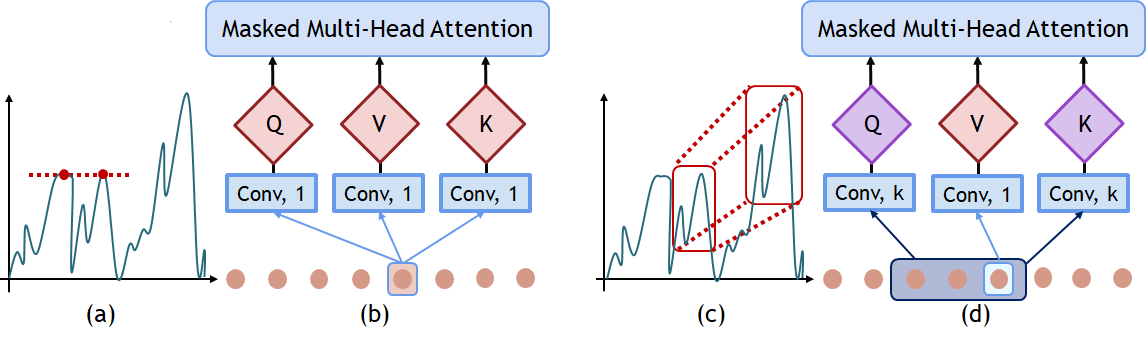}
  \caption{The canonical self-attention mechanism is similar to using convolution with kernel size equal to one, as shown in (b) \cite{li2019enhancing}. It is only able to capture point-wise similarity, as shown in (a). Local context is captured using convolution with kernel size greater than one proposed by Li et al. and depicted in (d). Awareness of the local context helps capture feature similarities more accurately based on shape matching, as shown in (c).}
  \label{fig:Localcontext}
\end{figure}

The patterns in time-series data may evolve significantly with time due to different events, like holidays or extreme weather. Therefore, it may be beneficial to capture information from the surrounding time points to determine whether the observed point is an anomaly, changing point, or a part of the pattern. Such behavior is captured using the causal convolutional self-attention mechanism, as shown in Figure \ref{fig:Localcontext}. 

The causal convolutional self-attention mechanism ensures that the current position does not have access to future information. The convolution operation with a kernel size of more than one captures the local context information in the query and key vectors generated from the input. Value vectors are generated using a kernel size equal to one. With this mechanism, more accurate forecasting is performed. 

\subsection{Simply Attend and Diagnose (SAnD)}
Clinical data such as Intensive Care Unit (ICU) measurements comprise of multi-variate, time-series observations coming from sensor measurements, test results, and subjective assessments. Song et al. introduced Transformers to predict various variables of clinical interest from the MIMIC-III benchmark dataset \cite{johnson2016mimic}, named Simply Attend and Diagnose (SAnD) \cite{song2017attend}. Preprocessed data is passed to an input embedding layer, which uses 1-D convolution for mapping inputs into $d$ dimensional vectors ($d$ $>$ number of variables in the multi-variate time-series data). After the addition of hard-coded positional encoding to the embedded data, it is passed through the attention module. The multi-head attention layer uses restricted self-attention to introduce causality, i.e., perform computations using information earlier than the current time. The output of the stacked attention modules passes on to the `dense interpolation layer'. This layer, together with positional encoding, is used to capture the temporal structure of clinical data. A linear layer and softmax/sigmoid are used as the final layers for classification. The proposed Transformer model was evaluated on all MIMIC-III benchmark tasks and was reported to perform better than RNNs.

\subsection{Traffic Transformer}
\begin{figure}[ht]
  \centering
  \includegraphics[width=1\textwidth]{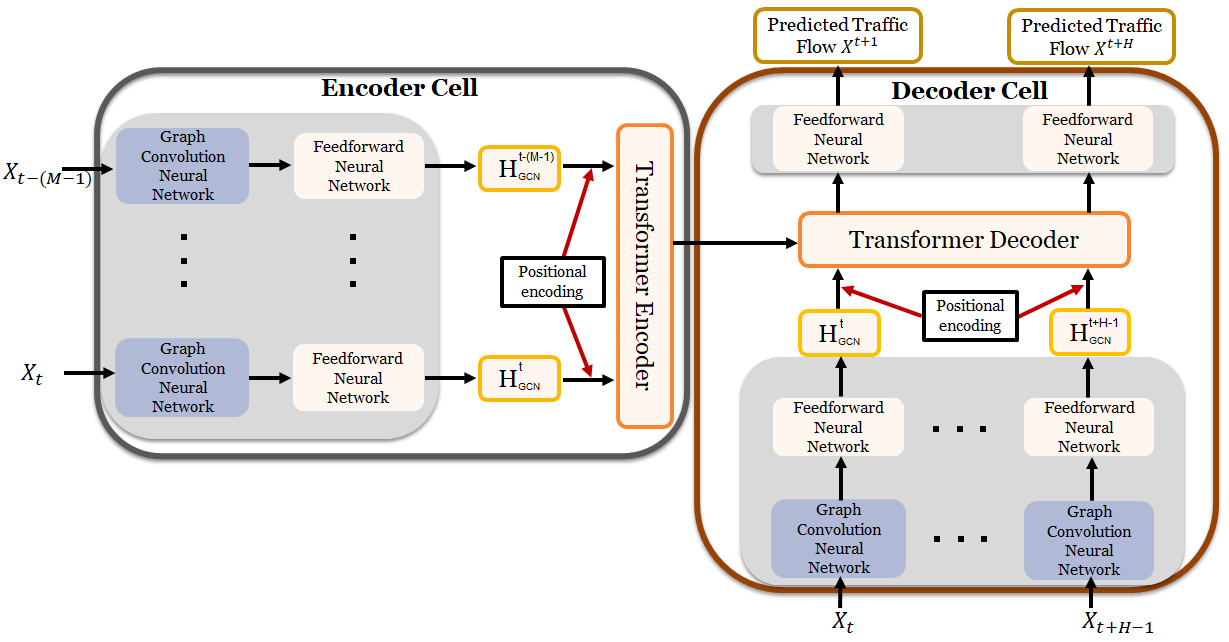}
  \caption{The Traffic Transformer introduced in \cite{cai2020traffic} is presented. The encoder comprises of Graph Neural Network for capturing the spatial dependencies in the traffic data, followed by a feed-forward layer. The output is fed into the Transformer encoder to capture the temporal dependencies. The output of the encoder is fed into the Transformer decoder, which has a similar composition to the encoder. Position information is included using either addition or similarity-based combination.}
  \label{fig:TrafficTransformer}
\end{figure}

Traffic forecasting predicts future traffic, given a sequence of historical traffic observations like speed, density, and volume detected by sensors on a road network. In this case, $M$ previous time steps in a sequence are used to predict $H$ future time steps. It is important to encode the continuity and periodicity of time-series data and capture the spatiotemporal dependencies in traffic forecasting. The Traffic Transformer \cite{cai2020traffic} has been built based on two existing networks, the first being a Graph Neural Network \cite{scarselli2008graph, yang2019aligraph} followed by the Transformer as shown in Figure \ref{fig:TrafficTransformer}. The Transformer models the temporal dependencies, and the Graph Neural Network is used to model spatial dependencies.

The output of the Graph Neural Network forms the input for the Transformer after the incorporation of positional encoding. There are two methods for adding the sequence information, (1) positional encoding vectors are added to the input vectors, and (2) attention weights for positional encoding vectors are calculated using the dot product. The positional attention weights are used to tweak the attention weights for the input vectors to the Transformer. Four strategies listed below for encoding temporal information of traffic data are introduced. Different combinations of these strategies can be used for various problems or dataset types.

\begin{itemize}
    \item[a)]Continuity of time-series data:
    \begin{itemize}
        \item[i.] Relative position encoding: This strategy encodes the relative position of a time step in the window of the source-target sequence regardless of the position of that time step in the entire time series. Hence, the same time step might be assigned with a different position embedding depending on its position in a sequence pair. Position encoding is done using sine and cosine functions.
        \item[ii.] Global position encoding: The entire sequence is encoded using sine and cosine functions. In this way, both local and global positions of time-series data are captured.
    \end{itemize}
    \item[b)] Periodicity of time-series data:
      \begin{itemize}
        \item[i.] Periodic position encoding: A mechanism to encode the daily and weekly periodicity of data is introduced. Daily periodicity is captured by encoding two hundred and eighty-eight ($\frac{24\times 60}{5}$) positions. Weekly periodicity is captured by encoding seven positions (number of days in a week).
        \item[ii.] Time-series segments: This is done by concatenating the daily and weekly data segments to the ‘M’ recent time steps.
     \end{itemize}
\end{itemize}

The Traffic Transformer was tested with two real-world benchmark datasets, METR-LA \cite{jagadish2014big} and the California Transportation Agencies Performance Measurement System \href{https://pems.dot.ca.gov/}{([PeMS])}.

\subsection{Self-Attention for Raw Optical Satellite Time-Series Classification}
Vegetation life cycle events can be used as distinct temporal signals to identify types of vegetation. These temporal signals contain relevant features for differentiating various kinds of vegetation. The classification of vegetation type \cite{russwurm2020self} from raw optical satellite images is done using Transformers. The performance of the Transformer architecture for vegetation classification, compared to LSTM-RNN \cite{hochreiter1997long}, MS-ResNet \cite{wang2017time}, DuPLO \cite{interdonato2019duplo}, TempCNN \cite{pelletier2019temporal}, and random forest methods, is better for raw data. However, for pre-processed data, the performance of all methods is quite similar.

\subsection{Deep Transformer Models for Time-Series Forecasting: The Influenza Prevalence Case}
Transformers have also been used to predict cases of influenza \cite{wu2020deep} using the data of weekly count of flu cases in a particular area. A single week's prediction is made using ten previous weeks' data as the input in the first experiment. The Transformer architecture performed better in terms of ‘root mean square error’ (RMSE) compared to ARIMA, LSTMs, and sequence-to-sequence models with attention. In the second experiment, the Transformer architecture was tested with multivariate time-series data by introducing ‘week number’ as a time-indexed feature and including the first and second-order differences of the time-series data as two explicit numerical features. However, this did not show significant improvement in the results. In the third experiment, Time-delay embedding (TDE) is introduced and formed by embedding each scalar input $x_t$ into a d-dimensional time-delay space as in Equation \ref{eq:tde}.

\begin{equation}
\text{TDE}_{d,\tau} x(t) = \left( x_t, x_{t-1},\cdots,x_{t-(d-1)\tau} \right)
\label{eq:tde}
\end{equation}
Time-delay embedding of different dimensions $d$, from $2$ till $32$, are formed with $\tau=1$. Out of all the experiments, the RMSE value reached a minimum with the dimension of $8$, which conforms to similar results of an independent study on clinical data \cite{sugihara1990nonlinear}.

\section{Best Practices for Training Time-Series Transformers}
\label{section:best-practices}
The Transformer architecture is becoming increasingly popular and leading many researchers to seek ways to optimize these networks for various applications. The approaches include adapting the architecture for use in a specific problem domain, adapting training techniques, hyperparameter optimization, inference methods, hardware adaptation, and others. In this section, we discuss best practices when training Transformers for time-series analysis. 

\subsection{Training Transformers}
The Transformers may not be easy to train from scratch for a beginner. The original Transformer architecture \cite{Vaswani2017} utilized many different strategies to stabilize the gradients during training for deeper networks. The use of residual connections allows for training a deeper network. Later, layer normalization operations were added alongside an adaptive optimizer (Adam) to provide different learning rates for different parameters. Like most other deep learning models, Transformers are also sensitive to the learning rate. Given an optimal learning rate, Transformers can converge faster than traditional sequence models. During the first few epochs, it is common to observe a performance drop. However, after a few epochs, the model will generally start to converge to better values. In the original implementation, the authors used a warm-up learning rate strategy that increases linearly for the first $N$ training steps and then decreases proportionally to the inverse square root of the step number, $\frac{1}{\sqrt{N}}$. 

\subsection{Implementing Transformers in Popular Frameworks}


Here, we provide an overview of popular frameworks for implementing and training Transformer models. These frameworks offer a user-friendly interface and support for custom model architectures, enabling researchers and developers to experiment with and adapt Transformers for time-series analysis and other tasks.

Hugging Face Transformers is an open-source library that provides pre-trained Transformer models and user-friendly APIs for a wide range of natural language processing tasks. Compatible with both PyTorch and TensorFlow, the library offers an extensive collection of pre-trained models and community-contributed models, making it an excellent starting point for researchers and developers. More details about the library can be found at the Hugging Face Transformers GitHub repository.

PyTorch Lightning is a lightweight wrapper for PyTorch designed to simplify deep learning research and development. It offers a structured approach to training deep learning models, including Transformer models, with minimal boilerplate code. PyTorch Lightning is compatible with the Hugging Face Transformers library, enabling users to leverage pre-trained models and fine-tune them for specific tasks. More information is available at the PyTorch Lightning GitHub repository. Other libraries in PyTorch for efficient large Transformer model training include Microsoft DeepSpeed and MosaicML Composer. 

TensorFlow is an open-source machine learning library developed by Google that supports building and training deep learning models, including Transformers. The library provides implementations of the original Transformer architecture and various pre-trained models that can be fine-tuned for specific tasks. TensorFlow also offers extensive documentation and community support for implementing custom Transformer models. The TensorFlow website offers more information.

\subsection{Improvements in the Transformer Architecture for Better Training}
\begin{figure}[ht]
  \centering
  \includegraphics[width=.5 \textwidth]{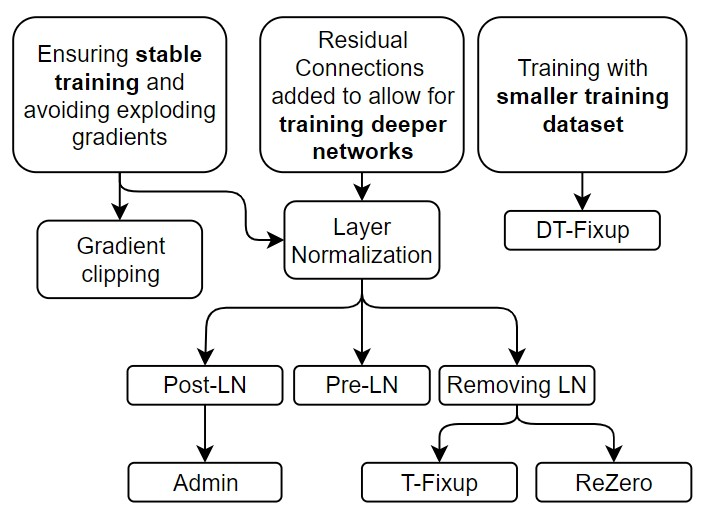}
  \caption{An overview of best practices for training Transformers. }
  \label{fig:bestpracfig2}
\end{figure}

Many advancements to the Transformer architecture have been proposed with the aim of resolving some of the problems related to achieving stable training with deeper architectures. Mainly optimizations have been made to balance residual dependencies by relocating layer normalization operations and finding better weight initialization techniques. These improvements have led to more stable training and, in some cases, eliminated the need for using some of the strategies proposed in the original architecture. Figure \ref{fig:bestpracfig2} provides an overview of some of the best practices for training Transformers and their respective reasons.

The original Transformer architecture can be referred to as post-layer normalization (post-LN), where the layer normalization is located outside the residual block \cite{preLN}. Post-LN converges much slower and requires a learning rate warm-up strategy \cite{preLN}. Xiong \emph{et al.} proposed a pre-layer normalization (pre-LN) Transformer to address this issue, showing that it could help gradients converge faster while requiring no warm-up. The Pre-LN Transformer can achieve this by controlling the gradient magnitudes and balancing the residual dependencies \cite{preLN}. Although the Pre-LN Transformer architecture does not require learning rate warm-up, it has inferior empirical performance as compared to the post-LN Transformer architecture.

\begin{figure}[ht]
  \centering
  \includegraphics[width=.5 \textwidth]{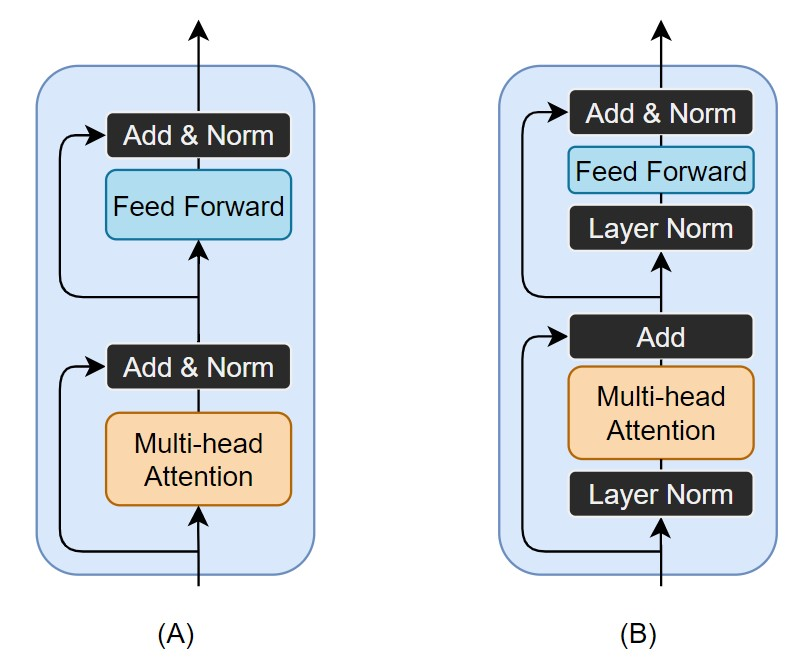}
  \caption{post-LN Transformer (A) vs. pre-LN Transformer (B)}
  \label{fig:bestpracfig3}
\end{figure}

Liu \emph{et al.} proposed an adaptive model initialization (Admin) to benefit from the ease of training of the post-LN Transformer while achieving the performance of the pre-LN Transformer \cite{Admin}. \emph{Adaptive model initialization} is a technique used in other areas of machine learning to initialize the model such that the dependencies between the input and output variables are better captured \cite{Admin}. This initialization technique helps the model learn the relationships between the variables more effectively and improves performance. Another option is to remove the layer normalization altogether. The \emph{ReZero} approach replaces the layer normalization operation with a trainable parameter, $\alpha$, which is initialized to $0$ in each residual layer \cite{pmlr-v161-bachlechner21a}. Consequently, the entire network is initialized to compute the identity function, with a gradual and adaptive introduction of the contributions of the self-attention and MLP layers. The residual dependencies seem to balance well with Admin or ReZero for training deeper Transformer models with better generalization performance.

\subsection{Practical Issues for Training Transformers}

\subsubsection{Large Model Size}
After selecting the Transformer model architecture, the challenge becomes dealing with the large model size. It may seem that smaller models would train faster than larger ones. However, this does not always hold true. For example, Li \emph{et al.} \cite{DBLP:journals/corr/abs-2002-11794} demonstrated that in some instances, training large models and then compressing the results leads to better performance. The \emph{lottery ticket hypothesis} \cite{frankle2018lottery} reveals that \emph{pruning} can be applied to reduce the model size. Also, \emph{quantization} to a lower precision allows for a smaller model. However, there are inevitable trade-offs when using a larger model after pruning/quantization versus a smaller one. The most relevant is ensuring that the training dataset is large to avoid overfitting. Using a small dataset can lead to poor generalization. In general, when trying to train a Transformer model, we recommend using large models instead of the more conventional approach of starting with smaller models and then adding layers.

\subsubsection{Training with Small Datasets}
We have seen several solutions that allow the training of deeper Transformer models with improved performance as compared to the vanilla Transformer architecture. Training these deep Transformer models from scratch requires large datasets, making training with a small dataset challenging. Small datasets require pre-trained models and small batch sizes to perform well. However, these two requirements make training additional Transformer layers more difficult. Using a small batch size causes the variance of the updates to be larger. Even with large batch size, the model may typically generalize poorly. In some cases, better initialization techniques can optimize the model making it perform well on the smaller datasets. The use of Xavier initialization is the most common scheme for Transformers \cite{Xavier}. Recently, T-Fixup has been shown to outperform Xavier \cite{T-fixup}. The motivation for the T-Fixup scheme, introduced by Huang \emph{et al.}, is as follows. It was found that when training a Transformer without learning rate warm-up, the variance in the Adam optimizer was amplified by a large initial learning rate thereby leading to large updates at the beginning of training \cite{T-fixup}. Hence, the learning rate warm-up requirement comes from the Adam optimizer's instability combined with gradient vanishing through layer normalization. To resolve this, the T-Fixup method was proposed in which a new weight initialization scheme provides  theoretical guarantees that keep model updates bounded and removes both warm-up and layer normalization \cite{T-fixup}. Following the work of T-Fixup, a data-dependent initialization technique (known as DT-Fixup) was developed \cite{xu-etal-2021-optimizing}. DT-Fixup allows for training deeper Transformer models with small datasets, given that the correct optimization procedures are followed. 

\subsubsection{Other Strategies to Consider}
\textbf{Batch Size.} Popel \emph{et al.} found that the optimal batch size depends on the complexity of the model \cite{Popel_2018}. They considered two classes of models, one a ``base'' and one a ``big.'' For the base model, a larger batch size (up to 4,500) performed well, whereas a different set of parameters yielded superior results for the big model.The big model needed a minimum batch size (1,450 in their experiments) before it started to converge. While batch size is almost entirely empirically selected, a large minimum should be used with Transformer models.

\noindent\textbf{Learning Rate.} Considerable research has been done in evaluating the effect of learning rate on model performance \cite{zeiler2012adadelta,smith2018disciplined,behera2006adaptive} and how they relate to each other. A study conducted by Popel \emph{et al.} demonstrated that small learning rates tend to have slower convergence, whereas learning rates that are too high may lead to non-convergence \cite{Popel_2018}. 

\noindent\textbf{Gradient Clipping.} Many implementations of Transformers use gradient clipping during the training. This process tends to avoid exploding gradients and potential divergence if the number of steps is too large. Other methods of gradient clipping, such as choosing a step size proportional to the batch size, are not recommended for Transformers as these can lead to slower convergence \cite{Popel_2018}.

\section{Conclusion and Future Trends}
\label{section:summary-conclusion}
In conclusion, the Transformer architecture has proven to be a powerful tool for tackling time-series tasks, offering an efficient alternative to RNNs, LSTMs, and GRUs, while also overcoming their limitations. To effectively handle time-series data, modifications to the original Transformer architecture have been proposed. Various best practices for training Transformers have been developed and many open-source frameworks are available for efficiently training large Transformer models.
Robustness \cite{waqas2022exploring}, failure detection \cite{ahmed2022failure}, and multimodal learning \cite{waqas2023multimodal} are some of the future trends in deep learning. Moving forward, the development of robust, self-aware Transformer architectures using uncertainty estimation in time-series prediction is an open challenge currently being pursued by the research community. The availability of large datasets having multiple modalities such as images, videos, and text, combined with time-series data can lead to the development of Transformer-based foundation models capable of learning indiscernible and subtle features that may lead to unprecedented discoveries for the time-series tasks.

\section{Acknowledgement}
\label{section:summary-conclusion}
This work was partly supported by the National Science Foundation Awards ECCS-1903466, OAC-2008690, and OAC-2234836. 

\section{Data Availability Statement}
\label{section:data}
The manuscript has no associated data.

\section{Funding and/or Conflicts of interests/Competing interests}
\label{section:conflict}
There are no conflicts of interest or competing interests.

\bibliographystyle{unsrt}  
\bibliography{references}

\end{document}